\documentclass[conference]{IEEEtran}
\IEEEoverridecommandlockouts
\usepackage{amsmath,amssymb,amsfonts}
\usepackage{algorithmic}
\usepackage{bm}
\usepackage{graphicx}
\usepackage{hyperref}
\usepackage{textcomp}
\usepackage{tikz}
\usepackage{xcolor}
\def\BibTeX{{\rm B\kern-.05em{\sc i\kern-.025em b}\kern-.08em
    T\kern-.1667em\lower.7ex\hbox{E}\kern-.125emX}}
    
\usepackage[symbol]{footmisc}
\usepackage{siunitx}
\newcommand{\figref}[1]{\figurename~\ref{#1}}

\renewcommand{\vec}[1]{\boldsymbol{\mathrm{#1}}}
\newcommand{\mat}[1]{\vec{#1}}
\DeclareMathOperator*{\argmax}{arg\,max}

\newcommand{\norm}[1]{\left\lVert#1\right\rVert}
\usepackage[OT2,T1]{fontenc}
\DeclareSymbolFont{cyrletters}{OT2}{wncyr}{m}{n}
\DeclareMathSymbol{\Sha}{\mathalpha}{cyrletters}{"58}

\usepackage[
backend=biber,
bibencoding=utf8,
style=ieee,
firstinits=true,
doi=false,
isbn=false,
url=false,
sorting=none
]{biblatex}
\AtNextBibliography{\footnotesize}

\newcommand\copyrighttext{\footnotesize \textcopyright~2018 IEEE. Personal use of this material is permitted. Permission from IEEE must be obtained for all other uses, in any current or future media, including reprinting/republishing this material for advertising or promotional purposes, creating new collective works, for resale or redistribution to servers or lists, or reuse of any copyrighted component of this work in other works.
	DOI: \href{https://ieeexplore.ieee.org/document/8547092}{10.1109/SDF.2018.8547092}
}%

\newcommand\copyrightnotice{%
	\begin{tikzpicture}[remember picture,overlay]
		\node[anchor=south] at (current page.south) {\fbox{\parbox{\dimexpr\textwidth-\fboxsep-\fboxrule\relax}{\copyrighttext}}};
	\end{tikzpicture}%
}

\begin{document}

\title{Motion Classification and Height Estimation of Pedestrians Using Sparse Radar Data}

\author{
\IEEEauthorblockN{Markus Horn\IEEEauthorrefmark{1}\IEEEauthorrefmark{2}, Ole Schumann\IEEEauthorrefmark{2}, Markus Hahn\IEEEauthorrefmark{2}, J\"urgen Dickmann\IEEEauthorrefmark{2} and Klaus Dietmayer\IEEEauthorrefmark{1}}
\IEEEauthorblockA{\IEEEauthorrefmark{1}Institute of Measurement, Control and Microtechnology, Ulm University\\
	Albert-Einstein-Allee 41, 89081 Ulm, Germany\\
	Email: \{markus.horn, klaus.dietmayer\}@uni-ulm.de}
\IEEEauthorblockA{\IEEEauthorrefmark{2}Daimler AG\\
	Wilhelm-Runge-Str. 11, 89081 Ulm, Germany\\
	Email: \{ole.schumann, markus.hahn, juergen.dickmann\}@daimler.com}
}

\maketitle

\begin{abstract}
A complete overview of the surrounding vehicle environment is important for driver assistance systems and highly autonomous driving.
Fusing results of multiple sensor types like camera, radar and lidar is crucial for increasing the robustness.
The detection and classification of objects like cars, bicycles or pedestrians has been analyzed in the past for many sensor types.
Beyond that, it is also helpful to refine these classes and distinguish for example between different pedestrian types or activities.
This task is usually performed on camera data, though recent developments are based on radar spectrograms.
However, for most automotive radar systems, it is only possible to obtain radar targets instead of the original spectrograms.
This work demonstrates that it is possible to estimate the body height of walking pedestrians using 2D radar targets.
Furthermore, different pedestrian motion types are classified.
\end{abstract}


\section{Introduction}
\label{sec:introduction}

\copyrightnotice

It is crucial for driver assistance systems and highly autonomous driving to create a detailed vehicle environment model.
Using multiple redundant sensors helps to increase the overall system accuracy and robustness.
This makes it necessary to perform the same task with all sensor types.
We focus on two different aspects: body height estimation of walking pedestrians and pedestrian motion classification using radar data.

A classification of dynamic objects like cars, bicycles or pedestrians based on radar data is explained for example by Schumann et al. \cite{schumann2017comparison}.
In order to refine the environment model, it is helpful to distinguish between different subclasses of pedestrian motion like \texttt{walk}, \texttt{skateboard} or \texttt{wheelchair}, and between children or adults.
This information could be used to act more cautious in potentially dangerous situations, for example around children or skateboarders.
Lei et al. \cite{lei2005target}, Kim et al. \cite{kim2009human} and Du et al. \cite{du2014noise} introduce such classifiers based on micro-Doppler signatures.
Micro-Doppler signatures are superpositions of Doppler modulations from a single object, visible in the time-Doppler spectrogram.
They show flapping wings of birds, rotating helicopter blades or swinging arms and legs of pedestrians \cite{chen2014radar}.

However, it is not possible with most automotive radar systems to extract the original spectrograms containing the full micro-Doppler signatures from the sensor and fuse them with other radar sensors.
Instead, most sensors provide sparse radar targets extracted for example with a constant false-alarm rate (CFAR) detector.
These radar targets represent reflection points of an object, which means multiple targets are detected on each object, depending on radar resolution, object location and size.
Although the Doppler velocities of these targets do not reflect the original micro-Doppler signature, they show similar oscillations.
Our focus is set on developing additional preprocessing steps for compensating the data sparsity in order to detect the original micro-Doppler signature.

The first part of this work introduces a body height estimation of walking pedestrians based on radar targets.
The available radar targets are two-dimensional, which means they provide only position and Doppler velocity over ground.
Therefore, the proposed body height estimation is based on the relation between pedestrian velocity, stride length and height.
This connection is described by the Boulic-Thalmann model for human walking \cite{boulic1990global}.
Since we are not able to measure stride parameters directly, they are extracted from radar target data.
A drawback is the limitation on walking pedestrians.
The second part of this work introduces a pedestrian motion classification for different activities.
Since radar targets still show oscillations of different moving body parts, it is possible to predict motion types based on the associated signatures.
We train a classifier for the motion types \texttt{walk}, \texttt{run}, \texttt{jump}, \texttt{crutches}, \texttt{skateboard} and \texttt{wheelchair}.

The remainder of this work is structured as follows:
In section~\ref{sec:related_work}, related work is discussed.
Section~\ref{sec:methods} describes the available data and briefly explains the applied machine learning techniques.
The feature extraction for body height estimation and motion classification is introduced in section~\ref{sec:height_estimation} and section~\ref{sec:classification}.
Finally, section~\ref{sec:results} explains the experiments and discusses the results before conclusions are drawn in section~\ref{sec:conclusion}.

\section{Related Work}
\label{sec:related_work}

Most previous classifiers for pedestrian motion are based on radar spectrograms instead of radar targets \cite{lei2005target}, \cite{kim2009human}, \cite{du2014noise}.
These spectrums contain the full micro-Doppler signatures.
The most similar data foundation is used by Kim et al. \cite{kim2009human},~\cite{kim2016human}.
They propose classifiers based on support vector machines and convolutional neural networks that can distinguish motion types like \texttt{walk}, \texttt{run}, \texttt{crawl} or \texttt{sit} on spectrograms of \SI{3}{\second} time windows.
In contrast, this work tries to achieve similar results on sparse radar targets instead of the original spectrogram, which makes it necessary to apply additional preprocessing steps.
Height estimation of pedestrians is primary based on camera data, whereas only a few approaches deal with body height estimation on radar data.
Dorp et al. \cite{vandorp2003human} extract the cycle parameters of walking pedestrians from radar spectrograms using trajectories given in the Boulic-Thalmann model.
Estimating the body height using 2D radar target data is a novel approach, which makes it possible to fuse the results with existing algorithms based on camera data.

\section{Methods}
\label{sec:methods}

\subsection{Data}
\label{sec:data}
The data are provided by four $\SI{77}{\GHz}$ radars integrated in the left and right front bumper of the test vehicle.
They are mounted with overlapping field of views and cover an angle of over $200^\circ$ in front of the vehicle.
The radars can detect targets within a distance range of $\SIrange{0.25}{250}{\m}$ with an accuracy of $\SI{0.2}{\m}$ for point targets.
Speed measurements are given with an accuracy of $\SI{0.1}{\km\per\hour}$.

\begin{figure}[!h]
	\centerline{\includegraphics[width=0.90\linewidth]{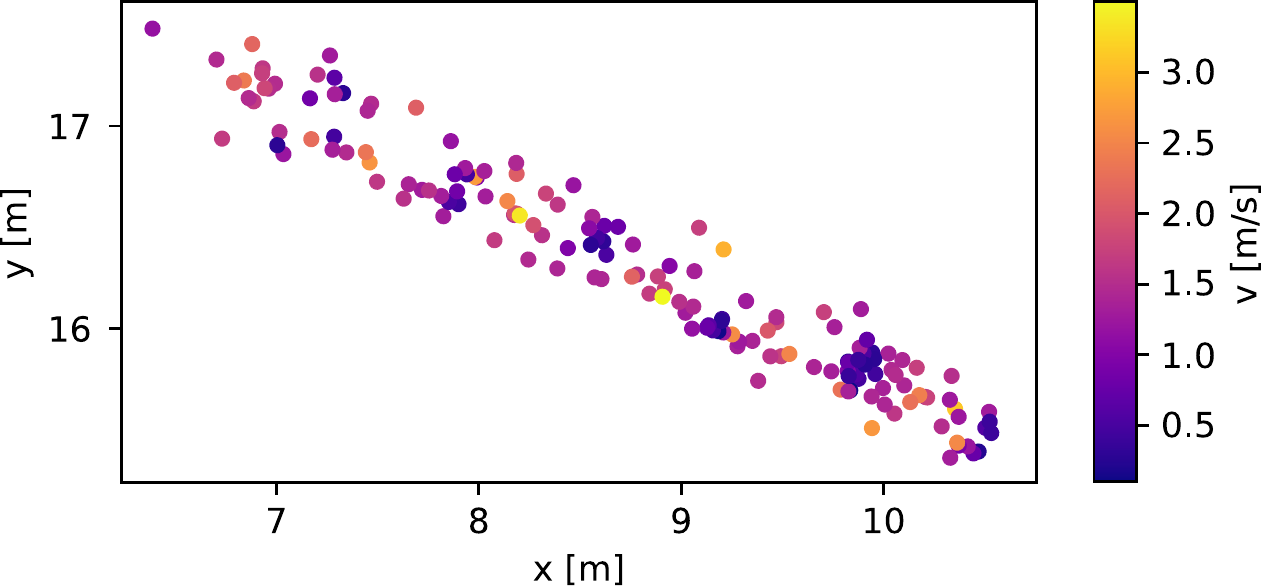}}
	\caption{Radar target data of a walking pedestrian with the Doppler velocity as point color. An earth fixed global coordinate system is used in this representation.}
	\label{fig:data}
\end{figure}

The radar interface provides preprocessed data in form of detected radar targets, extracted from spectrogram data with a constant false-alarm rate (CFAR) detector.
Single objects like pedestrians are detected as multiple targets, depending on radar resolution, object location and size.
Each target measurement $i$ consists of range and angle $(r_i, \phi_i)$ in sensor coordinates, ego-motion compensated radial velocity over ground $v_i$ and the corresponding timestamp $t_i$.
The position in sensor coordinates is transformed to a position $(x_i, y_i)$ in an earth-fixed coordinate system with the initial vehicle pose as origin using sensor calibration and vehicle ego-motion.
Timestamps are non-uniformly distributed with an average sample time of $\SI{18}{\ms}$.
For processing, we collect targets for a fixed time period $T = \SI{3}{\s}$ and bulk process the data afterwards.
The number of targets within a time window is defined as $N$.
\figref{fig:data} shows radar targets of a walking pedestrian with the Doppler velocities as point colors.
Dark blobs represent positions of feet on the ground with Doppler velocities of almost zero.

\subsection{Random Forests}
\label{sec:random_forests}

The proposed classifier is based on random forests, originally introduced by Breiman~et~al.~\cite{breiman2001random}.
They support regression as well as classification, which are used for body height estimation and motion classification, respectively.
We use the implementation provided by the free Python machine learning library Scikit-learn \cite{pedregosa2011scikit}. 
Our contribution is the development of features that help to distinguish various motion types and describe the relation between radar target data of a walking pedestrian and its body height.
The main focus is set on compensating the missing information due to the sparse data representation.

\subsection{Sparse Dictionary Learning}

Sparse dictionary learning and sparse coding for training and prediction, respectively, are unsupervised machine learning techniques.
The objective is to use training data to create a dictionary $\mat{D}$ with few base vectors called atoms.
This dictionary can be used to approximate every training vector $\vec{x}_i$ as linear combination $\vec{\tilde{x}}_i = \mat{D} \vec{\alpha}_i$ of atoms.
Sparse coding is the process of finding linear coefficients $\vec{\alpha}_i$ for given vector $\vec{x}_i$ and dictionary $\mat{D}$.
In the context of image classification, $\vec{x}$ is a vectorized image and atoms represent base images.
A dictionary is found by minimizing the cost function
\begin{equation} \label{eq:dict_min}
	\min_{\mat{D}, \mat{\alpha}} \frac{1}{n} \sum_{i=1}^{n} \left( \frac{1}{2} \norm{\vec{x}_i - \mat{D} \vec{\alpha}_{i}}_{2}^{2} + \lambda \norm{\vec{\alpha}_{i}}_1 \right)
\end{equation}
with training data $\lbrack \vec{x}_1, \dots, \vec{x}_n \rbrack$, coefficient matrix $\mat{\alpha} = \lbrack \vec{\alpha}_1, \dots, \vec{\alpha}_n \rbrack$ and hyperparameter $\lambda$.
The hyperparameter $\lambda$ balances the tradeoff between reconstruction error and sparsity.
The cost function is optimized by alternating optimization of dictionary atoms and coefficients.
A detailed description of an online dictionary learning algorithm is given by Mairal~et~al.~\cite{mairal2009online}.
Once a sparse dictionary is generated, the coefficients for new data can be found with the sparse coding algorithm in \cite{mairal2009online}.

\section{Height Estimation}
\label{sec:height_estimation}

The proposed height estimation is based on the correlation between stride length $l_s$ in m, pedestrian velocity $v_{ped}$ in \si{\metre\per\second} and body height $h$ in m.
This connection is described by the Boulic-Thalmann model for human walking \cite{boulic1990global}.
By combining the equations for height of thigh $h_t$, relative stride length $\tilde{l}_s = \frac{l_s}{h_t}$ and relative velocity $\tilde{v}_{ped} = \frac{v_{ped}}{h_t}$
\begin{eqnarray} \label{eq:boulic}
	h_t &=& 0.53 \cdot h\\
	\tilde{l}_s &=& 1.346 \sqrt{\tilde{v}_{ped}},
\end{eqnarray}
we can describe the body height of an average human with 
\begin{equation} \label{eq:height_model}
	h = \frac{l_s^2}{1.346^2 \cdot 0.53 \cdot v_{ped} }.
\end{equation}
This is only a rough estimate, generalized for a wide range of body heights and designed for simulating human walking.
Our experiments have shown that the Boulic-Thalmann model cannot be used directly for estimating the body height of walking pedestrians.
In contrast, random forest regression is able to describe more complex relationships.

The major regression parameters are stride length $l_s$ and pedestrian velocity $v_{ped}$.
In the next section, we describe the trajectory estimation, which is used to obtain $v_{ped}$.
The stride length is extracted from the Doppler velocity $v_i$ over tangential distance $d_i$ in Frenet coordinates.
We can calculate the stride length by determining the cycle frequency of these data, as described in section~\ref{sec:stride_length}.

\subsection{Frenet Transformation}
\label{sec:frenet_transformation}

\begin{figure}[!h]
	\centerline{\includegraphics[width=0.8\linewidth]{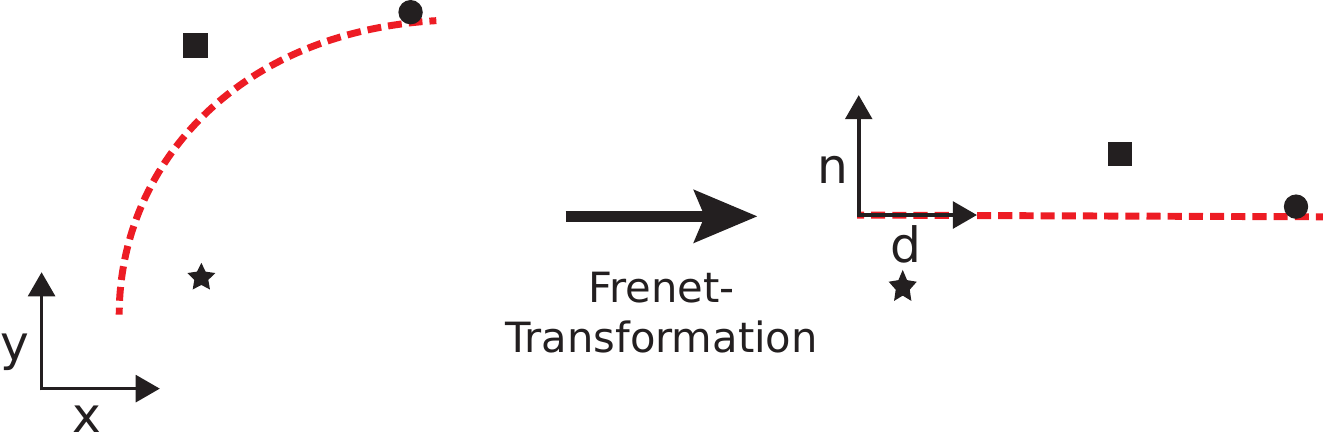}}
	\caption{Frenet coordinate transformation example for three points (black) and a curved reference path (red).}
	\label{fig:class:frenet_coordinates}
\end{figure}

Before extracting the stride length, the radar target positions are transformed into Frenet coordinates.
As explained in section~\ref{sec:data}, radar targets are given in an earth-fixed coordinate system with position $(x_i, y_i)$.
The Frenet transformation calculates tangential distance $d_i$ and orthogonal distance $n_i$ of each target in reference to a given path \cite{werling2010optimal}.
\figref{fig:class:frenet_coordinates} shows an example for the Frenet coordinate transformation with a curved reference path.
The Frenet transformation makes it possible to use the Doppler velocities as signal over tangential distance for extracting the stride length.

Since we only consider small time frames of $\SI{3}{\s}$, it is reasonable to approximate the pedestrian movement using a linear trajectory with constant velocity $v_{ped}$.
The RANSAC scheme \cite{fischler1987random} is used for estimating the trajectory.
In each iteration, the algorithm picks two random targets to calculate $x(t)$ and $y(t)$.
All remaining targets are checked for inliers, i.e.\ points within a fixed range around the calculated trajectory.
After a fixed number of iterations, the largest inlier set is used to estimate the trajectory with an overdetermined linear equation system.
It is important to note that the calculation only relies on the time $t_i$ and on the $(x_i, y_i)$ position of the targets, whereas the Doppler velocity is not used for trajectory estimation.
The Frenet transformation is finally realized by orthogonal projection of each target on the trajectory path.
Even though simple rotation and translation could be used for straight trajectory paths, we do not limit ourselves to this special case and use the Frenet transformation as the more general approach.

\subsection{Stride Length}
\label{sec:stride_length}

This section describes the stride length extraction from Doppler velocity $v_i$ over non-uniform sampled tangential distance $d_i$ with $N$ samples.
We calculate the step length $l_{step}$ by inverting the step frequency $f_{step}$ in $\si{\per\metre}$ obtained from the data.
The stride length is $l_s = 2 \cdot l_{step}$.
The algorithm is separated into three steps:

\begin{enumerate}
	\item \textbf{Moving Average Filtering}\\
	The first step is to create uniformly sampled data of the Doppler velocity $\tilde{v}_j$ over distance $\tilde{d}_j = d_0 + j \cdot \Delta d$ with sample distance $\Delta d = \SI{0.1}{\metre}$ and $j=0,\dots,M-1$.
	Furthermore, the original signal is low-pass filtered to suppress the influence of undesired high frequencies in the following FFT.
	This is achieved with an exponential moving average filter for non-uniformly sampled data \cite{eckner2015algorithms}.
	In contrast to an ordinary moving average filter, the weights are based on a Gaussian normal distribution.
	This filter function is used because of its better frequency characteristic compared to a rectangle window.
	The new velocity values are calculated as
	\begin{equation}
		\label{eq:height:lowpass}
		\tilde{v}_j = \frac{\sum_{i = 0}^{N - 1} v_i \cdot F(d_i | \tilde{d}_j)}{\sum_{i = 0}^{N - 1} F(d_i | \tilde{d}_j)} \quad \text{for} \ j = 0, \dots, M-1
	\end{equation}
	where $F(d|\mu)$ denotes the Gaussian normal distributed filter function $F(d|\mu) = \exp \left\lbrace - (d - \mu)^2 / (2 \sigma^2) \right\rbrace$ with fixed variance $\sigma$.
	The parameter $\sigma$ should not be selected too high, since the filter can also cause attenuation of lower frequencies, where the step frequency is extracted.
	Experiments have shown that $\sigma = \SI{0.03}{\metre}$ has almost no negative influence on the spectrum within the interesting range.
	
	\item \textbf{Windowing and Zero-Padding}\\
	Since the processed data section is not cyclic in most cases, a window function is applied.
	The Hann window \cite{harris1978use} is used to reduce the influence of side lobes resulting from the subsequent FFT.
	Additionally, the signal is zero-padded for increasing the frequency accuracy.
	
	\item \textbf{Fast Fourier Transform and Frequency Extraction}\\
	After creating uniformly sampled data, the FFT is used to generate the frequency spectrum $\vec{V}_f$ with discrete frequencies $f$ in $\si{\per\metre}$.
	The step frequency $f_{step}$ of the input data is obtained from the spectrum as
	\begin{equation}
		f_{step} = \argmax_{f \in \left[ f_{min}, f_{max} \right]} V_f.
	\end{equation}
	The stride length is calculated from the step frequency as $l_s = 2 / f_{step}$.
	The frequency range is set to $f_{min} = \SI{0.8}{\per\metre}$ and $f_{max} = \SI{2.5}{\per\metre}$.
\end{enumerate}

\subsection{Regression Features}
\label{sec:regression}

Random forests for regression are built from multiple regression trees, similar to the classification equivalent.
In our case, the regression features are only based on pedestrian velocity $v = v_{ped}$ and stride length $l = l_s$.
The pedestrian velocity $v$ is calculated from the linear trajectory in section~\ref{sec:frenet_transformation}.
The algorithm described in the previous section is used to obtain the stride length $l$ for a walking pedestrian.
Polynomial and rational features are added to represent the complex relation between body height and features.
The final feature vector used for training the regression forest for body height estimation is
\begin{equation} \label{eq:height:features}
\left[ v, \, l,\, v l,\, v^2 l,\, v l^2,\, \frac{l}{v},\,  \frac{l}{v^2},\, \frac{l^2}{v} \right].
\end{equation}

\section{Motion Classification}
\label{sec:classification}

The proposed features for pedestrian motion classification can be divided into two categories: target based features and grid based features.
Target based features are extracted directly from the set of radar targets and require no additional preprocessing, cf.\ section~\ref{sec:signal_features}.
Grid based features are created after transforming the data into an image-like grid representation.
This structure makes it possible to apply pattern recognition methods from image processing: In addition to a feature vector created from a normalized histogram of oriented gradients (HOG), we also obtain features using class-specific dictionary learning.
All feature types are eventually combined to a single feature vector and used to train a classification forest.

\subsection{Signal Features}
\label{sec:signal_features}
The first feature set contains simple statistic parameters of the Doppler velocity data without additional preprocessing steps.
Statistic moments \cite{miller1966probability} can describe basic characteristics of the data.
We use the central moments
\begin{equation}
	\mu_k = \frac{1}{N - 1} \sum_{i = 1}^{N} (v_i - \mu)^k \quad \text{for} \ k = 2, \dots, 4
\end{equation}
and the first central absolute moment 
\begin{equation}
	\bar{\mu}_k = \frac{1}{N - 1} \sum_{i = 1}^{N} \left| v_i - \mu \right|\ .
\end{equation}
The mean Doppler velocity $\mu$ is not used, since it varies depending on the angle between radar and pedestrian direction.
A pedestrian walking diagonal to the radar direction is detected with attenuated Doppler velocities, since the radial velocity part is smaller.

\subsection{Grid Transformation}
\label{sec:grid_transformation}

\begin{figure}[!h]
	\centerline{\includegraphics[width=0.9\linewidth]{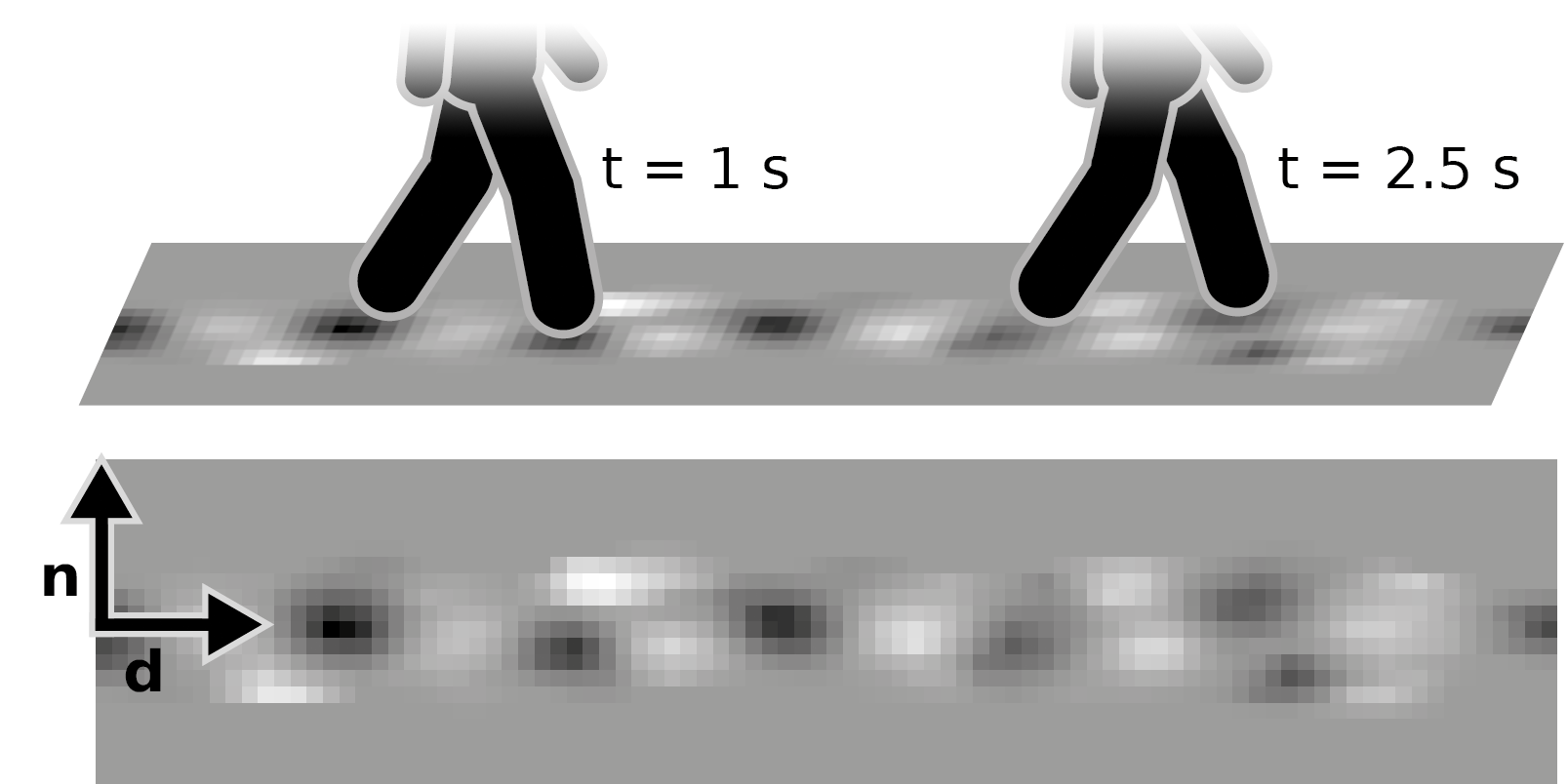}}
	\caption{Grid transformation example of a walking pedestrian. Dark regions indicate low velocities, bright regions high velocities.}
	\label{fig:class:grid_transform}
\end{figure}

In order to enable uniform data processing independent of the pedestrian path, the targets are first transformed into Frenet coordinates as explained in section~\ref{sec:frenet_transformation}.
For the grid transformation, tangential and orthogonal distance are divided into a 2D grid.
The grid cell values contain the deviation of the associated mean Doppler velocity to the pedestrian velocity~$v_{ped}$, empty grid cells are filled with zero.
This means, the value zero indicates either that the mean Doppler velocity for this cell is equal to $v_{ped}$ or that the cell is empty.
Before creating the discrete grid structure, the data are filtered with a Gaussian in order to prevent aliasing effects.
This leads to a structure of connected blobs with high or low velocities.
After generating the grid, it is normalized to $[-1, 1]$, where zero still represents $v_{ped}$.
\figref{fig:class:grid_transform} shows a normalized grid of a walking pedestrian.

\subsection{Image Features}
\label{sec:image_features}

The grid representation makes it possible to use features from pattern recognition techniques for images.
We use histograms of oriented gradients (HOG) features~\cite{dalal2005histograms}.
First, the gradients in both image directions are extracted with simple 1D masks $[-1, 0, 1]$ and combined to a 2D gradient for each grid cell.
Afterwards, the unsigned gradient orientation $\tilde{\theta}_G = \theta_G \mod{180^\circ}$ in $[0^\circ, 180^\circ)$ is split into evenly sized bins.
The unsigned gradient angle basically ignores the `sign' of the gradient direction.
Each grid cell votes with its magnitude for the two closest gradient bins.
To reduce aliasing, the values are interpolated bilinearly between neighboring bin centers.
The final values of the bins are normalized to a sum of one and used as classification features.

\subsection{Dictionary Features}
\label{sec:dictionary_features}

Dictionary learning can be used for classification by generating a separate dictionary for each class during training.
This is called supervised class-specific dictionary learning~\cite{wang2012supervised}.
In our case, the training data for dictionary learning are vectorized 2D FFT images of the grid representations.
The FFT is used in order to obtain a shift invariant representation, since the time windows can be arbitrary located within the motion cycle.

For prediction, a FFT image with unknown class is approximated with each class-specific dictionary using sparse coding.
With the resulting coefficients, the reconstruction for each class is generated with the corresponding atoms.
The class associated with the reconstruction causing the lowest reconstruction error is selected as prediction.
All other dictionaries are less suited for representing the input image.
The error is calculated as mean squared error between original image and reconstruction.
Finally, the prediction is converted to a feature vector for the random forest using one-hot encoding.
The final one-hot encoded feature vector contains an element for each class, with the corresponding entry for the predicted class set to 1 and the remaining entries set to 0.

\section{Experimental Results}
\label{sec:results}

\subsection{Experiments}

The height estimation is evaluated with 56 walking test subjects with a body height between \SI{1.53}{\metre} and \SI{2.00}{\metre}.
With over \SI{73}{\percent} between \SI{1.70}{\metre} and \SI{1.90}{\metre}, the subject heights are unevenly distributed.
The data of ten test subjects performing the actions \texttt{walk}, \texttt{run}, \texttt{jump}, \texttt{crutches} and \texttt{skateboard} for approximately \SI{45}{\second} each are available.
A single test subject performed the \texttt{wheelchair} action for \SI{66}{\second}.
The pedestrians were only moving back and forth parallel to the radar direction with a distance up to \SI{25}{\metre}.
The recorded radar targets were clustered and labeled manually.
For creating test samples, the time windows of \SI{3}{\second} were selected overlapping with an offset of \SI{1}{\second} to increase the amount of available samples.

\begin{table}[htbp]
\caption{Motion classification sample count per class.}
\vspace*{-0.6cm}
\begin{center}
\begin{tabular}{|c|c|c|c|c|c|c|}
\hline
\textbf{class} & walk & run & jump & crutches & skateboard & wheelchair \\
\hline
\textbf{samples} & 493 & 203 & 235 & 726 & 364 & 45 \\
\hline
\end{tabular}
\label{tab:motion_samples}
\end{center}
\end{table}

This results in 2135 samples for body height estimation and 2169 samples for motion classification.
Table~\ref{tab:motion_samples} shows the motion sample distribution.
Most samples are available for the classes \texttt{walk} and \texttt{crutches}, whereas only 45 samples are available for the \texttt{wheelchair} class.
Both body height estimation and motion classification were evaluated with \mbox{5-fold} cross-validation.
In order to prevent overlaps between training and test set, the data were grouped by test subjects.
Only the \texttt{wheelchair} data were divided by recordings, since only a single test subject was available.
The macro-averaged F1-score and the mean absolute error (MAE) are used as performance indicators for motion classification and height estimation respectively.

The random forest parameters were optimized with a parameter sweep, which motivated the following design choices.
The Gini index and mean squared error were selected as split criteria for classification and regression.
A performance saturation was reached at 100 trees for height estimation and 50 trees for motion classification.
The tree depth of the classification was not limited, whereas the maximum regression tree depth was set to five.
Furthermore, the predictions were tested with other window lengths than \SI{3}{\second}.
Reducing the length does decrease the performance, whereas using a longer window time only improves the results slightly.

\subsection{Height Estimation}

First, the Boulic-Thalmann based model equation \eqref{eq:height_model} was used to calculate the body height with the extracted stride parameters.
This leads to a MAE of \SI{19.4}{\centi\metre} with a standard deviation of \SI{25.3}{\centi\metre}.
The results indicate that the model equation is not suitable for estimating the body height.
On the one hand, this is a result of the model generalization over a wide range of body heights.
In their original publication, they already mention large variations in their averaged values.
On the other hand, the stride parameter extraction is probably biased, which can lead to a large height estimation error.

\begin{figure}[!h]
	\centerline{\includegraphics[width=0.75\linewidth]{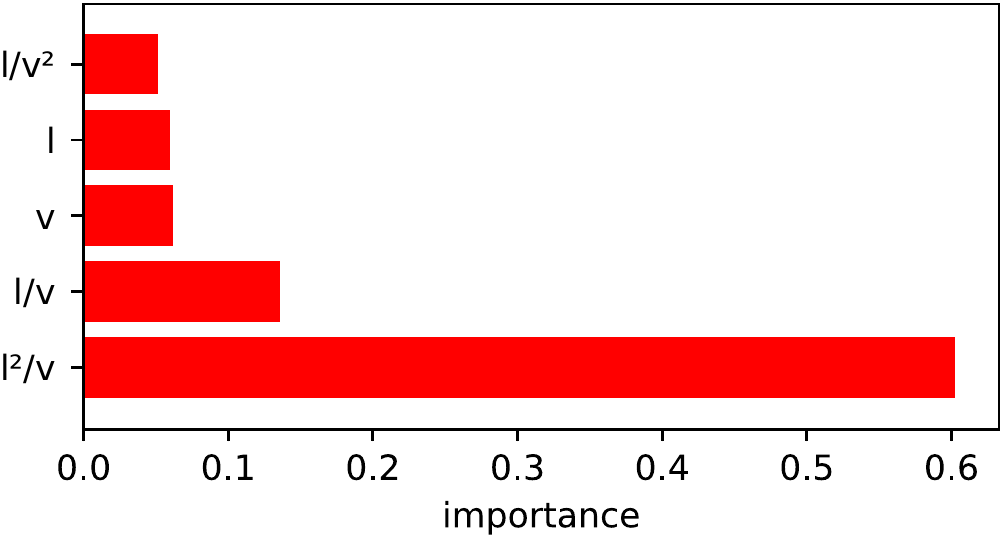}}
	\caption{Importance of the most relevant regression forest features for height estimation.}
	\label{fig:height_feature_importance}
\end{figure}

The performance is significantly higher, if a random forest for regression is trained with the feature set from equation \eqref{eq:height:features}.
The overall MAE is \SI{6.3}{\centi\metre} with a standard deviation of \SI{8.1}{\centi\metre}.
\figref{fig:height_feature_importance} shows the normalized feature importances.
The importance values are calculated from the influence of each feature on the Gini index during forest construction \cite{friedman2009elements}.
By far the most important feature is the factor $l^2 /v$, which is the same factor used in the Boulic-Thalmann based model equation \eqref{eq:height_model}.
This indicates that the basic structures of model equation and forest estimator are similar.
Nevertheless, the forest achieves better results than the model equation.
This is a result of the additional features used in the forest.
Furthermore, the forest is able to learn the relationship between factors and body height from training data, which helps to compensate a possible bias in the stride parameter extraction.

\begin{figure}[!h]
	\centerline{\includegraphics[width=0.78\linewidth]{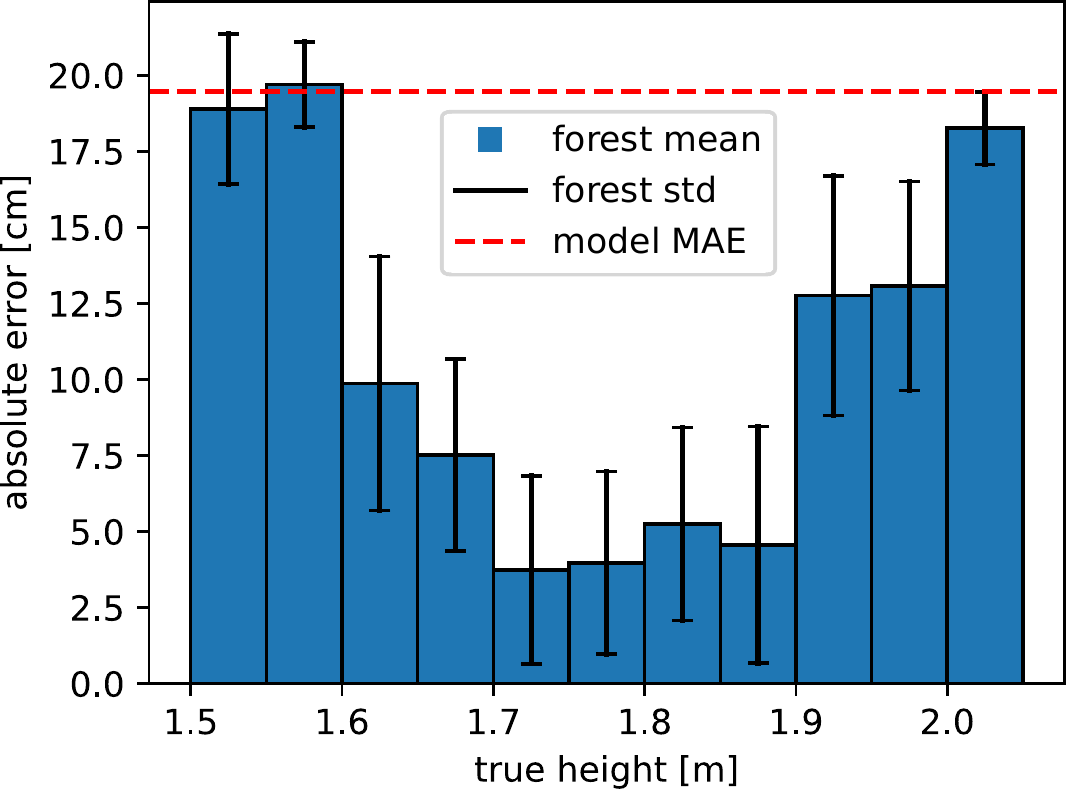}}
	\caption{
		Absolute error average (blue) and standard deviation (black) of the random forest based body height estimation for \SI{5}{\centi\metre} bins.
		The red line shows the mean absolute error resulting from the Boulic-Thalmann model equation.
	}
	\label{fig:height_error}
\end{figure}

\figref{fig:height_error} shows the overall MAE for \SI{5}{\centi\metre} height bins and the MAE of the Boulic-Thalmann based model as comparison.
Especially between \SI{1.70}{\metre} and \SI{1.90}{\metre}, where \SI{73}{\percent} of all test subjects are located, the MAE is below \SI{5}{\centi\metre}.
The estimation fails to generalize for body heights below \SI{1.60}{\metre} and above \SI{1.90}{\metre}.
This is a result of the uneven test subject height distribution.
Only two test subjects were smaller than \SI{1.60}{\metre} and only five test subjects were larger than \SI{1.90}{\metre}.

\subsection{Motion Classification}

\begin{figure}[!h]
	\centerline{\includegraphics[width=0.75\linewidth]{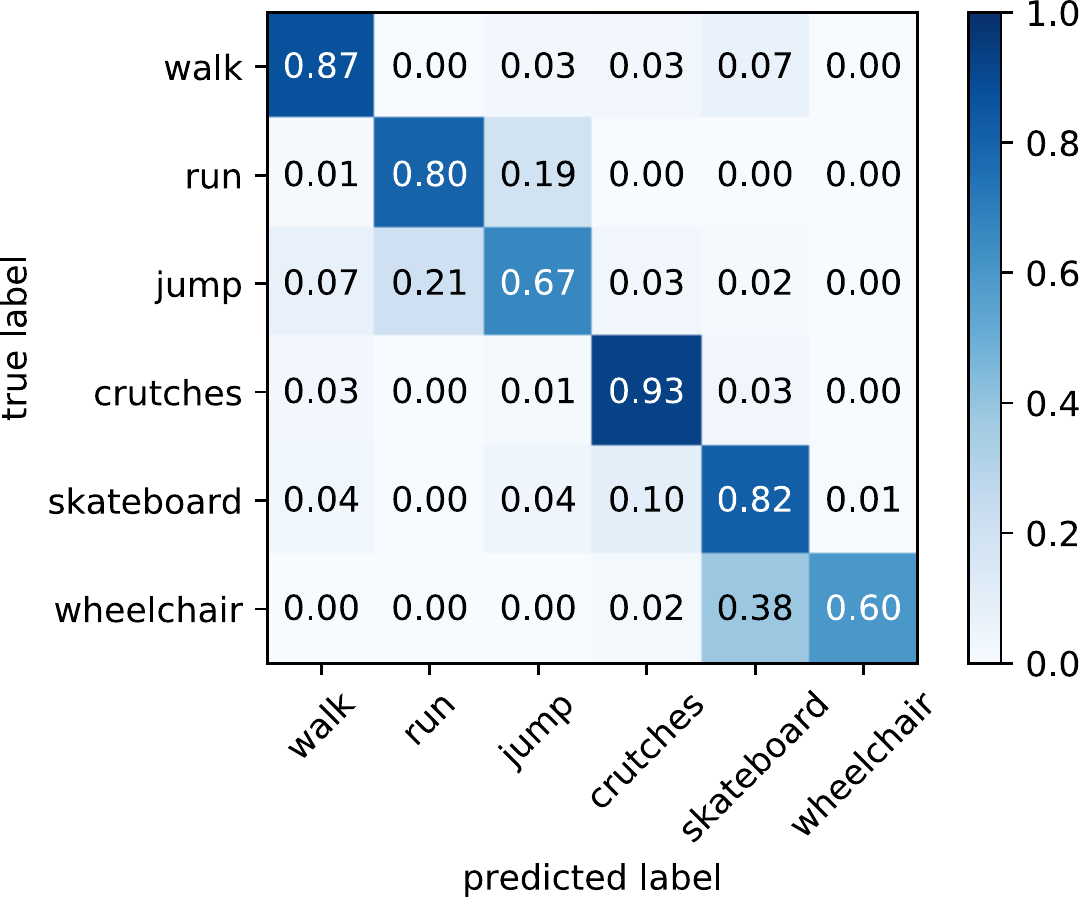}}
	\caption{Cross-validation matrix of the motion classification.}
	\label{fig:motion_cmat}
\end{figure}

\figref{fig:motion_cmat} shows the results of the motion classification.
The cross-validation matrix is normalized over the rows of true labels, which means that the diagonal elements show the recall.
The macro-averaged F1-score over all classes is 0.799, precision and recall are 0.830 and 0.781.
Especially the motion types \texttt{walk} and \texttt{crutches} are classified accurately with an F1-score of 0.881 and 0.922.
The types \texttt{run} and \texttt{jump} are confused in some cases, since they have similar signatures.
Only the recall of 0.60 for \texttt{wheelchair} data is low.
One reason for this is the low number of samples for this class, with only \SI{2}{\percent} stemming from \texttt{wheelchair} data.

\section{Conclusion}
\label{sec:conclusion}

This work introduces body height estimation and motion classification for pedestrians based on sparse 2D radar data.
We trained a body height estimator based on a given model for the relationship to velocity and stride length.
In comparison to the original model, this random forest based estimator leads to improved results for the entire height range of test subjects.
Although it is limited to walking pedestrians, it provides a useful approach to estimate the body height from 2D radar target data.
The body height is estimated with a mean absolute error of \SI{6.3}{\centi\metre}.
The high MAE of pedestrians smaller than \SI{1.60}{\metre} and larger than \SI{1.90}{\metre} is a result of the uneven height distribution of the test subjects.
Thus, it is crucial for future developments to acquire additional data from test subjects with a greater height range, e.g.\ from children and adolescents.

Our focus for motion classification was set on the necessary preprocessing steps for compensating the data sparsity.
This was achieved by applying the Frenet transformation for generating a specially developed grid representation of radar targets based on position and Doppler velocity.
The classification features were extracted with state-of-the-art pattern recognition techniques.
The final classifier has a macro-averaged F1-score of 0.799.
High performance is reported for the classes \texttt{walk} and \texttt{crutches} whereas pedestrians in \texttt{wheelchairs} are only recognized with \SI{60}{\percent} recall.
For comparison, Kim et al. \cite{kim2009human} achieve a higher performance with an F-score of 0.90.
However, they rely on the full spectrogram instead of CFAR level targets, which makes it impossible to apply their algorithm on data provided by automotive radar sensors as used in this work.
The performance can be improved in future developments e.g.\ by adding additional features based on the radar cross-section.
For further research, it is also necessary to obtain data in a real-world scenario with randomly selected test subjects.
Although the data only contain parallel movement to the radar, previous works like Kim et al. \cite{kim2009human} have shown that additional diagonal data only decrease their performance slightly.
Further possible applications of the proposed methods include observation of targets on water or through walls, similar to \cite{kim2016classification} and \cite{zenaldin2016radar}.

\printbibliography

\end{document}